\documentclass{article}
\usepackage[utf8]{inputenc}
\usepackage[english]{babel}

\usepackage{biblatex}
\usepackage{csquotes}
\usepackage{hyperref}
\usepackage{authblk}
\usepackage{setspace}
\usepackage[margin=1.25in]{geometry}
\usepackage{graphicx}
\graphicspath{ {./figures/} }
\usepackage{subcaption}
\usepackage{amsmath}

\title{CycleGAN with three different unpaired datasets}

\author[]{Tadem Sai Pavan}

\affil[]{SMST,Indian Institute Of Technology ,Kharagpur,India.}
\date{}

\begin{document}

\maketitle

\begin{abstract}
The original publication\cite{zhu2020unpaired} "Unpaired Image-to-Image Translation via Cycle-Consistent Adversarial Networks" served as the inspiration for this implementation project. Researchers developed a novel method for doing image-to-image translations using an unpaired dataset in the original study. Despite the fact that the pix2pix model's findings are good, the matched dataset is frequently not available. In the absence of paired data, cycleGAN can therefore get over this issue by converting images to images. In order to lessen the difference between the images, they implemented cycle consistency loss.I evaluated CycleGAN with three different datasets, and this paper briefly discusses the findings and conclusions.

\end{abstract}


\section{Introduction}

Image translation has become one of the most important areas to focus on.Deep learning advancements aided in resolving this issue. GAN contains two neural networks that compete against each other (adversarial) to produce more accurate predictions.Consider a case of translating winter to summer and then bacck to winter. CycleGAN consists of two GANs, i.e., two generators and two discriminators. One generator transforms selfies into anime, and the other transforms anime images back to selfies. Discriminators check whether the images generated by generators are fake or real during training. This loop ensures that an image created by a generator is cycle consistent. It means that, consecutively, both generators on an image should yield a similar image. Shooting problems with pix2pix..

In case of image translation (supervised) there is nothing to worry about what kind of output to be generated ,but in case of unsupervised (unpaired data) it is more important to focus on the task for example orange  to tomato
translation ,there can be two possibilities orange is completely translated to tomato or it can be a just colour change .In unsupervised learning data sets  plays a very important role to generate the mapping function .The special thing about GANs is,they can create new objects and by taking a random input.

\section{Related Work}

\textbf{Image-to-Image Translation :} 
The concept behind image to image translation is not only learn from the input images but also learns the loss to train the mapping .CycleGAN learns the mapping from original image to generated image, along with that it,learns a loss function to train this input and output mapping.

\textbf{Generative adversarial Networks:}
GANs are proved to be expert at translating image to image.The same idea can be applied to video ,text translations.The reason behind the success of GANs is adversarial loss. This loss forces the created images to be indistinguishable from the input image.The adversarial loss learn mapping function such a way that the generated images can not be distinguished from the target.GANs takes random input from Gaussian noise and generates meaningful outputs.GANs are generator and discriminator which are implemented by using convolutional neural networks to perform feature extraction and mapping.Both generator and discriminator are learned through back propagation,So generator learn to produces better desired results and discriminator learn to not to become a fool.

\textbf{Unpaired Image-to-Image Translation:}
Initially the image to image translation is supervised i.e in presence of paired data set ,but here they proposed an unsupervised model which is basically image to image translation from unpaired images.This model has wider range of applications due to being unsupervised,for example in health care,the medical data is dry and expensive for paired data.

\begin{figure}[h!]
    \centering
    \includegraphics[width=10cm]{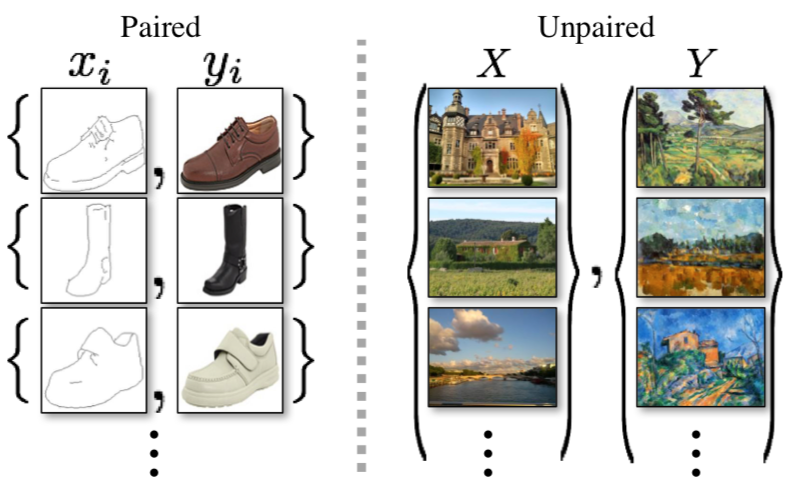}
    \caption{Paired and Unpaired Images}
    \label{fig:galaxy}
\end{figure}

\textbf{Neural style transfer:}
It is one among the methods to convert Image to Image translation,by this method we can generate style of domain x in domain y.
\textbf{Cycle consistency loss:}
In cycleGAN generator-G converts the original image i.e domain X to domain Y,and the generator-F tries to reconstruct that image from domain Y back to domain X,during this conversion it is important to keep follow up the loss every time to improve the reconstruction ,that responsibility is taken by cycle consistency loss.cycle consistency loss is added with the adversarial loss and back propagated,so that model become more clever at performing task. 

\section{Network Architecture}

\begin{figure}[h]
    \centering
    \includegraphics[width=0.8\textwidth]{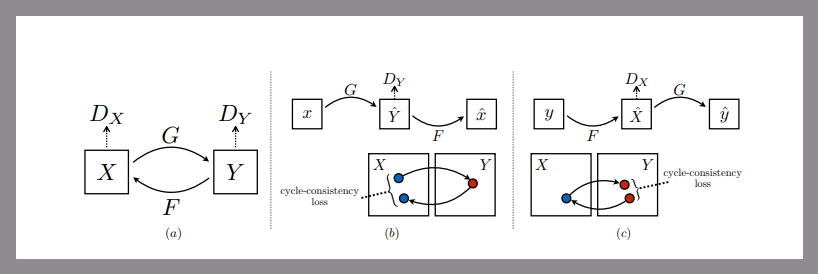}
    \caption{CycleGAN}
    \label{CycleGAN}
\end{figure}

\subsection{Model} 

In CycleGAN there are two discriminators  and two Generators.The model for the generative networks is taken from the Johnson\cite{johnson2016perceptual}.The results of that model are impressive for style transfer and super resolution.This architecture contains convolutional layers(three) , residual blocks(six), transpose convolutional layers(two) with stride of one and one convolutional layer which is going to map the features into RGB plane.The researcher's made slight modifications to the above mentioned network and designed the generators G and F,both are inverse of each other  and performs one to one map operation.The aim is to creating a mapping between domain X to domain Y and then back to domain X,consider $x_i$ belongs to X and $y_i$ belongs to Y,The mathematical view of the model is  $G : X \rightarrow Y$ and another translator $F : Y \rightarrow X $,So G and F are inverse of each other, and both the mappings are one to one .
\subsection{Generator}
Generator is a combination of encoder,decoder and residual layers.   Generator is responsible of creating a fake image from the input image. Generator is sub sectioned into three parts,encoder block,residual block and decoder block..
\subsubsection{Encoder}
Encoder plays a key role in feature extraction from the original image.
Encoder is designed with the help of convolutional layers.There are three convolutional layers in encoder,initial layer accepts input dimension of 3 and at output of third layer we have 256.Each convolutional layer is followed by instance normalization with the batch size of one.

\emph{IN-instance normalization,Conv-convolutional Layers.}
$\boldsymbol{Encoder: } Conv-64-IN,Conv-128-IN,,Conv-3-IN$
\subsubsection{Residual Block}
In generator encoder and decoder both are linked to gather by using residual blocks,deep learning models are having gradient vanishing issue,It is difficult to train them perfectly,So the results are not desired.But it can be solved by using residual blocks,which are going to learn the residual functions.
There are six resnet blocks each block contains two convolutional layers.
The overall architecture is as follows for generator, here Res stands for a block {Conv-IN}.The output of encoder is given to residual layers. 

$\boldsymbol{Residual blocks:} Res-256,Res-256,Res-256,Res-256,Res-256,Res-256$
\subsubsection{Decoder} 
The output of residual layers is taken by the decoders.Decoder is basically a up-sampler.It reconstructs the all the features into an image.This is constructed by two transpose convolutional layers takes 256 as input dimension and reduces to 32 as original.last layer is a convolutional layer takes 32 dimensions and converts into three dimensional image i.e RGB and this layer is followed by a tanh activation,here transpose convolutional layer is represented as TransposeConv.
$\boldsymbol{Decoder:} TransposeConv-128-IN,TransposeConv-64-IN,Conv-3-Tanh$

\subsection{Discriminator}
Discriminator is full of convolutional Layers( 5 layers) which are used to classify image patches of seventy by seventy sized  are fake or real.The role of patchGANs is to  double the number of channels and halves the size.This is repeated up to the point output converges to desired state. For discriminator the filter size is 3$*$3 ,It takes 3 dimensional input image and converts into 32,64,128,256 stage wise and back to dimension 3.In discriminator ReLU are Leaky with $\alpha$= 0.2.
$\boldsymbol{Discriminator:} Conv-32,Conv-64,Conv-128,conv-256,conv-3$

\section{Training Details:}
Adam\cite{kingma2017adam} is used as optimizer,it is one of the best performer out there,to decay the runtime average of gradients beta-1 is taken as 0.5 and to decay the square of the gradient beta-2is taken as 0.999, as initially the learning rate is set to 0.002 and the batch size is set to one i.e instance normalization which implies that batch normalization is not used .In all tasks the value of $\lambda$ is taken as 10.
\subsection{Generator Training}
The real image(x) is given to Generator-$G_{x}$ and it creates fake image fake\_.and that fake image given to the discriminator-$D\_{y}$ and it predicts the fake decision($D\_{y}$ fake\_decision) and by using that fake decision  mean square loss of $G\_{x}$ is estimated.The generated fake\_{y} is given to the Generator-G\_{y},and it tries to regenerate the original image(R\_{x}) with some loss.Now training Generator-G\_{y} it takes an original image(y) from domain Y and generates fake image i.e fake-x.This fake-x is passed to discriminator D\_{x} to and predicts fake decision(D\_{x}fake\_decision).Mean square loss of generator G\_{y} is calculated by using D\_{x}fake decision.
\textbf{Forward cyclic loss} is calculated from the reconstructed image(R\_{x}) and real image x.\textbf{Backward cyclic loss} is calculated from the reconstructed image (R\_{y}) and original image y.And total generator loss is sum of  G\_{x} loss, G\_{y} loss,forward cyclic loss and backward cyclic loss.This final loss is back propagated.In all cases Adam optimizer is used. 

\subsection{Discriminator Training}
The discriminator D\_{x} is trained with real image x and  fake\_{x} generates D$\_${x} real\_decision ,D\_{x} fake\_decision. 
 Decision loss of D$\_${x} is calculated by real and fake decisions and result is back propagated.Similarly discriminators D$\_${y}is trained with real image y , fake$\-${y} generates
 D$\_${y} real$\_$decision,D$\_${y} fake$\_$decision.By the help of real and fake decision,Decision loss of D$\_${y} is calculated by real and fake decisions and result is back propagated.

Both discriminators D\_{x} and D\_{y} are trained by using real image x,y and generates decision-x,decision-y.and both discriminators D\_{x} and D\_{y} are trained by using fake images x,y and generates fake-decision-x,fake-decision-y,sum of the real and fake losses of D\_{x} and D\_{y} are back propagated.

\subsection{Image buffer}
Generator and discriminator both are trained at a time,it is most important to take care of model not to change drastically for every simultaneous epochs.To avoid that\cite{shrivastava2017learning} discriminator is fed with the previously generated images,instead of just one image generated by the generator.In image pool we should store 50 recently generated images,If we train like this both generator and discriminator overfits and then mode collapse will going to occur,by doing this model oscillations\cite{goodfellow2017nips} and overfitting both can be reduced.

\section{Experiments:}
\textbf{Maps dataset:} Total number of samples in the data set are 1100 and  they partitioned  into trainA,trainB,testA,testB to train and test the model .This data set is collected from Kaggle\cite{Maps2satellite}. By this data set CycleGAN is trained up to 150 epochs with 0.02 learning rate and then applied linearly decay of learning rate up to 315 epochs.
\textbf{Vangogh2photo dataset:} It is a small data set also partitioned into four sets for training and testing purpose,trained up to 150 with the 0.002 learning rate and from 150 to 230 epochs with decayed learning rate,i.e learning rate becomes zero gradually.This data set is collected from kaggle \cite{Vangogh2photo}.
\textbf{Summer2winter dataset:} It  is used to show the season transfer.The entire data is partitioned into four parts and used for training and testing.
This is trained for 120 epochs with 0.002 learning rate and then from 120 to 230 epochs with linearly decay learning to zero.This data set is taken from the kaggle\cite{summer2winter} this images are normalized ro 256 * 256 pixels .Total summer training images are 1273 and winter images are 854.

\section{Objective functions}
There are two  loss functions in CycleGAN an adversarial loss and cycle consistency ,both are important and essential to bring good outputs.
Two Components to the CycleGAN objective function, an adversarial loss, and Cycle consistency loss.Both the generators tries to fool their respective discriminator.

The loss of mapping function $ G: X \rightarrow{} Y $is as follows.
\begin{equation}\label{eu_eqn1}
L_{GAN}(G, DY , X, Y ) = E_{ y,pdata(y)}[log DY (y)] + E_{x,pdata(x)}[log(1-DY (G(x))]
\end{equation}
The discriminator tries to maximise the above expression and generator tries to minimize the adversary Discriminator.The mathematical formulation is 
$ min_{G} max_{DY} L_{GAN}(G, DY , X, Y )$ .similarly $ F: Y to X $ mapping loss is as below
  
\begin{equation}\label{eu_eqn2}
L_{GAN}(F, DX , Y, X ) = E_{x,pdata(x)}[log DX(x)]+ E_{y,pdata(y)}log(1-DX (G(y))]
\end{equation}
similarly from $Y \rightarrow{}$ X the adversarial acts like  is  min$\_${F} max$\_${DX} L$\_${GAN}(F, DX, Y, X). Only Adversarial loss is not enough to produce good output(images),because we are training both Generators at a time we need to form a cyclic loss .
\begin{equation}\label{eu_eqn3}
L_{cyc}(G, F)=E_{x,pdata(x)}[||F(G(x))-x||_{1}]+ Ey,pdata(y)
[||G(F(y))-y||_{1}]. 
\end{equation}
The full objective function is formed by using above loss function together, and measuring the Cycle-consistency loss by using a hyper parameter $\lambda$ .
\begin{equation}\label{eu_eqn4}
L(G, F, DX, DY ) =L_{GAN}(G, DY , X, Y )
+ L_{GAN}(F, DX, Y, X)
+ \lambda L_{cyc}(G, F), 
\end{equation}

So the final aim is to solve optimise this equation arg min\_{G} max\_{Dx,DY} L(G, F, DX, DY ).

\section{Applications of CycleGANs}

we can apply cycleGAN to many areas of applications for example converting Selfie of person to anime,winter season to summer season ,Animal to animal,sketch to photo ,to convert low resolution images to high resolution,object transfer ,and photo enhancement.There are wide range of application in computer vision ,graphics,video games etc

\section{Results}
\textbf{NOTE-1} all Images are in the format of domain-X (original Image)$->$ domain-Y (generated)$->$ Reconstructed Image.

\begin{figure}[htbp]
\minipage{0.5\textwidth}
  \includegraphics[width=\linewidth]{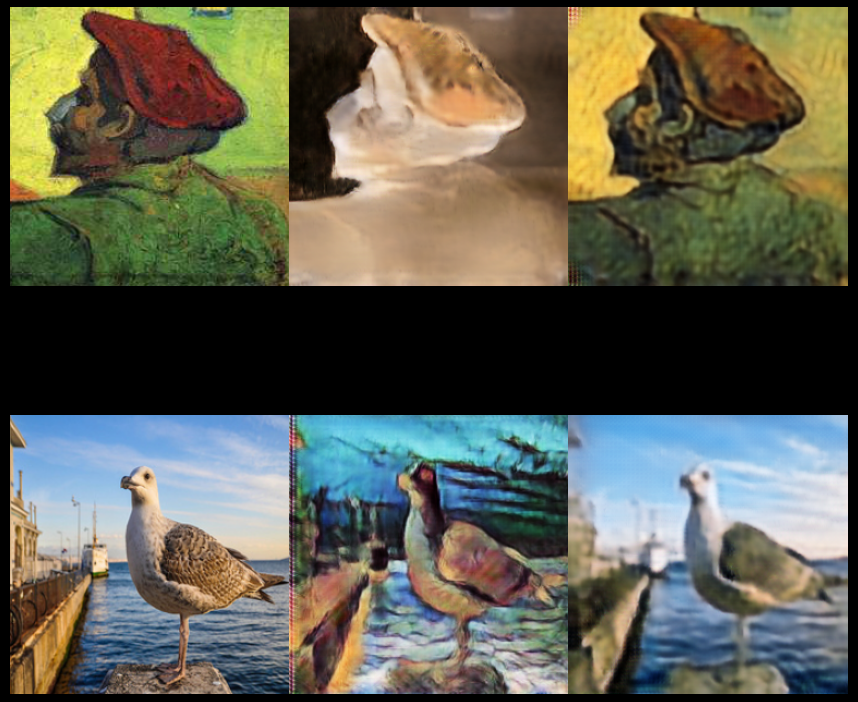}
  \caption{\textbf{Row-1:}Vangogh$>$Pictures$>$Vangogh and \textbf{Row-2:}Picture $>$ Vangogh$>$Picture}\label{fig:awesome_image1}
\endminipage\hfill
\minipage{0.5\textwidth}
  \includegraphics[width=\linewidth]{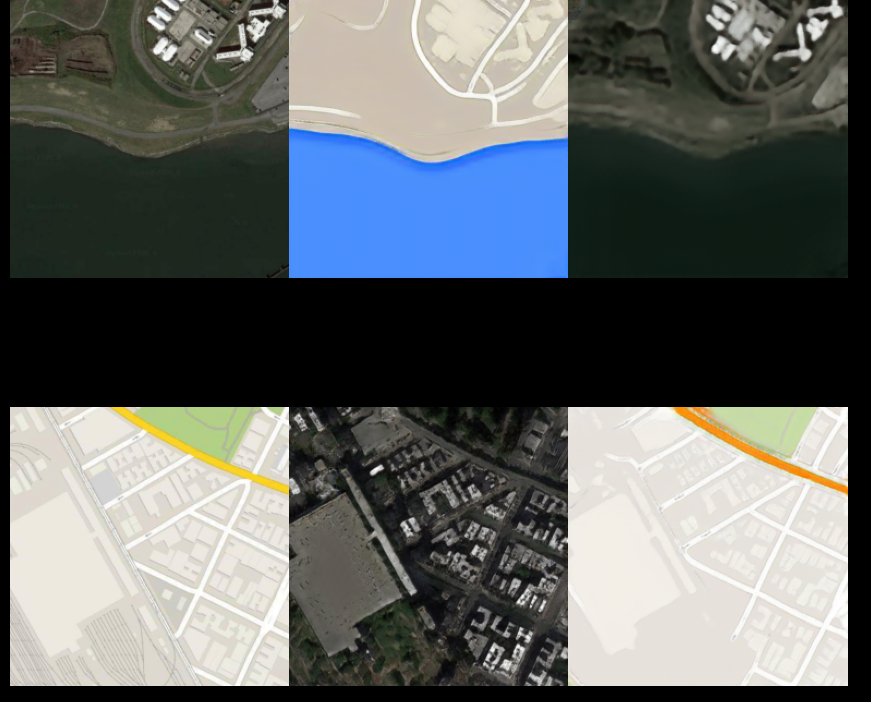}
  \caption{\textbf{Row-1:}Satellite$>$Aerial$>$Satellite and \textbf{Row-2:}Aerial$>$ Satellite$>$Aerial}\label{fig:awesome_image2}
\endminipage\hfill
\minipage{0.5\textwidth}%
  \includegraphics[width=\linewidth]{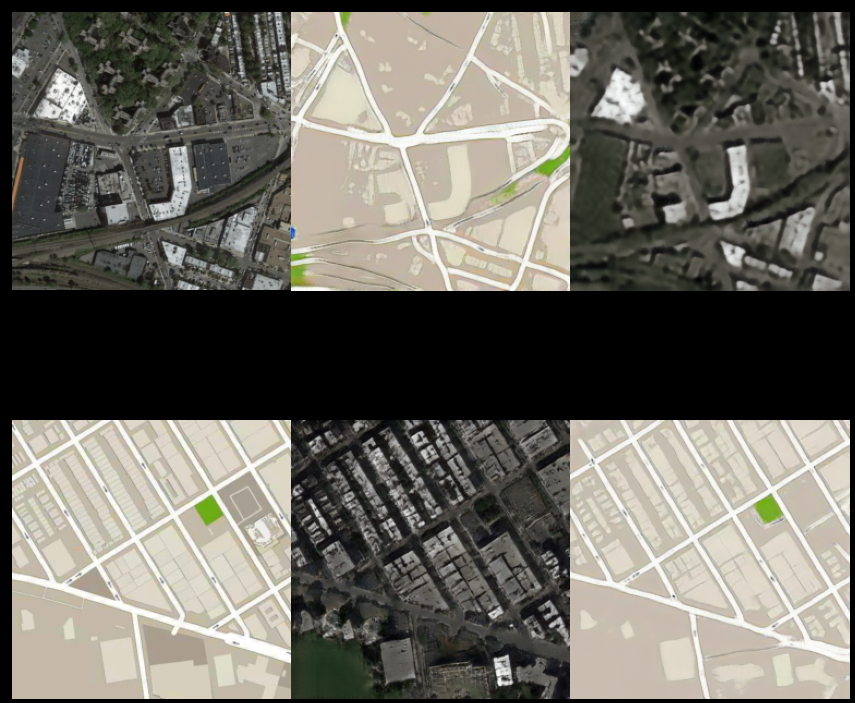}
  \caption{\textbf{Row-1:}Satellite$>$Aerial$>$Satellite and \textbf{Row-2:}Aerial $>$ Satellite $>$Aerial}\label{fig:awesome_image3}
\endminipage
\minipage{0.5\textwidth}
  \includegraphics[width=\linewidth]{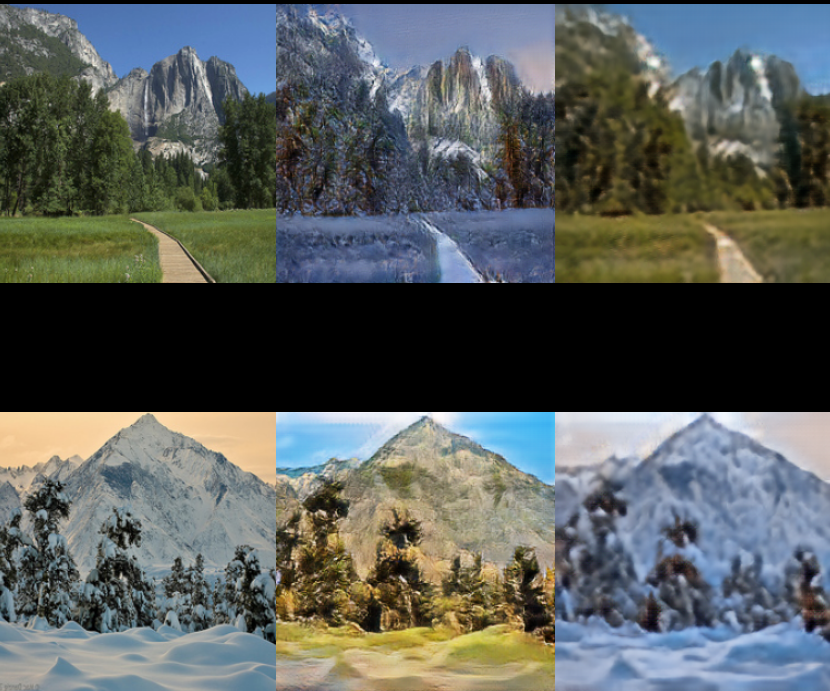}
  \caption{\textbf{Row-1:}Summer$>$Winter$>$Summer and \textbf{Row-2:}Winter$>$ Summer$>$Winter}\label{fig:awesome_image4}
\endminipage\hfill
\end{figure}
\subsection{Loss Curves}
The below loss curves are for summer to winter and winter to summer conversion task.There is no perfect measurement to show the performance of cycleGAN.But we can observe the model accuracy by the loss figures.
\begin{figure}[h!]
    \begin{subfigure}{0.6\textwidth}
        \includegraphics[width=0.9\textwidth, height=2in]{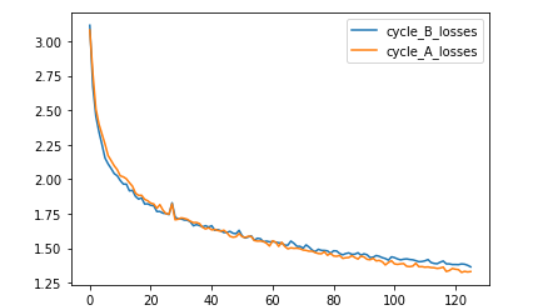}
        \caption{\label{fig:7a}}
    \end{subfigure}
    \begin{subfigure}{0.6\textwidth}
        \includegraphics[width=0.9\textwidth, height=2in]{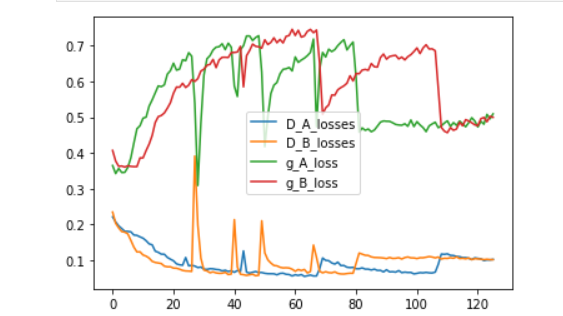}
        \caption{\label{fig:7b}}
    \end{subfigure}
    \caption{Loss plots for summer to winter vice versa.(\subref{fig:7a}) Cyclic losses. (\subref{fig:7b}) Generator and discriminator losses.}
    \label{fig:2}
    \maketitle
\end{figure}

\section{Limitations}
\begin{figure}[h!]
    \begin{subfigure}{0.6\textwidth}
        \includegraphics[width=0.9\textwidth, height=2in]{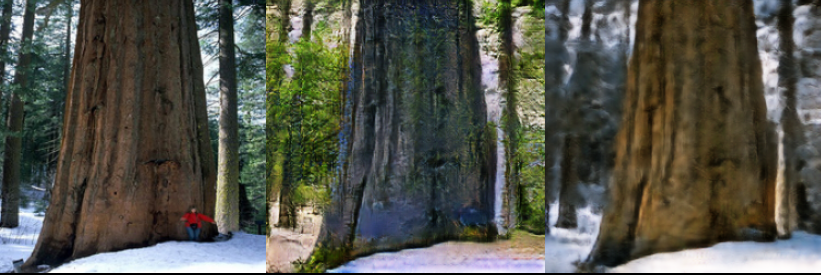}
        \caption{\label{fig:8a}}
    \end{subfigure}
    \begin{subfigure}{0.6\textwidth}
        \includegraphics[width=0.9\textwidth, height=2in]{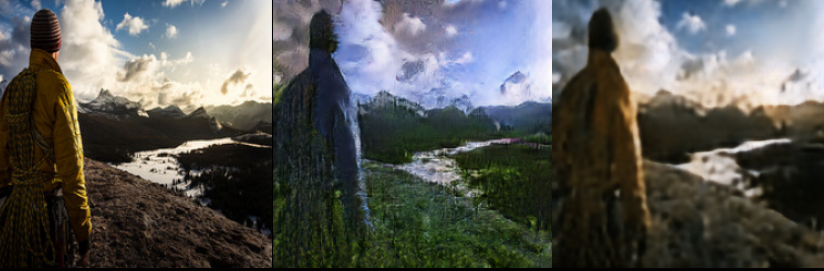}
        \caption{\label{fig:8b}}
    \end{subfigure}
    \caption{These are some figures which shows cycleGAN difficulties.     (\subref{fig:8a}) Summer to Winter conversion(Women is not there in generated Images). (\subref{fig:8b}) Statue is covered with the grass in generated image.}
    \label{fig:3}
    \maketitle
\end{figure}

Even though cycleGAN produces good results, there are some failure cases of summer to winter and winter to summer translations. In Fig-a, the woman in front of the tree is not present in the reconstructed image. In Fig-b, the generated image of a large statue is completely covered by grass, which is a major failure of cycleGAN. This observation is explained in the horse to zebra translation by the researchers in the original paper. Not only is there a resolution change from the original image to the generated and regenerated images, the clarity of the image is slightly reduced. 

\section{Conclusion}
The results are much more accurate in all three data sets used to test the cycleGAN, as demonstrated in the paper. few cases, the model is recreating with low resolution images. Sometimes background changes or colour changes are introduced. All the above results and limitations clearly show that CycleGAN still needs a good amount of research.   

\printbibliography

\end{document}